\documentclass[runningheads]{llncs}

\usepackage[mobile]{eccv}

\usepackage{eccvabbrv}

\usepackage{graphicx}
\usepackage{booktabs}
\usepackage{multirow}
\usepackage{makecell}
\usepackage{wrapfig}
\usepackage[dvipsnames, svgnames, x11names]{xcolor}
\usepackage{colortbl}
\usepackage[accsupp]{axessibility}  %

\usepackage[pagebackref,breaklinks,colorlinks,citecolor=eccvblue]{hyperref}

\usepackage{orcidlink}

\begin{document}

\title{Memorize-and-Generate: Towards Long-Term Consistency in Real-Time Video Generation
} 

\titlerunning{MAG}

\author{Tianrui Zhu\inst{1}\index{Zhu, Tianrui}$^{*}$ \and
Shiyi Zhang\inst{1}\index{Zhang, Shiyi}$^{*}$\and
Zhirui Sun\inst{1}\index{Sun, Zhirui}\and
Jingqi Tian\inst{1}\index{Tian, Jingqi}\and
Yansong Tang\inst{1}$^{\dagger}$}

\authorrunning{T.~Zhu et al.}

\institute{$^1$Tsinghua Shenzhen International Graduate School, Tsinghua University\\
\email{\{zhutr25, sy-zhang23, sunzr25, tjq25\}@mails.tsinghua.edu.cn}\\
\email{tang.yansong@sz.tsinghua.edu.cn}\\
\vspace{1pt}
$^{*}$Equal contribution \quad $^{\dagger}$Corresponding author
}

\maketitle
\vspace{-20pt}
\begin{abstract}
Frame-level autoregressive (frame-AR) models have achieved significant progress, enabling real-time video generation comparable to bidirectional diffusion models and serving as a foundation for interactive world models and game engines. However, current approaches in long video generation typically rely on window attention, which naively discards historical context outside the window, leading to catastrophic forgetting and scene inconsistency; conversely, retaining full history incurs prohibitive memory costs. To address this trade-off, we propose \textbf{Memorize-and-Generate (MAG)}, a framework that decouples memory compression and frame generation into distinct tasks. Specifically, we train a memory model to compress historical information into a compact KV cache, and a separate generator model to synthesize subsequent frames utilizing this compressed representation. Furthermore, we introduce \textbf{MAG-Bench} to strictly evaluate historical memory retention. Extensive experiments demonstrate that MAG achieves superior historical scene consistency while maintaining competitive performance on standard video generation benchmarks.
\end{abstract}

\vspace{-40pt}
\begin{figure}[h]
  \centering
  \includegraphics[width=122mm]{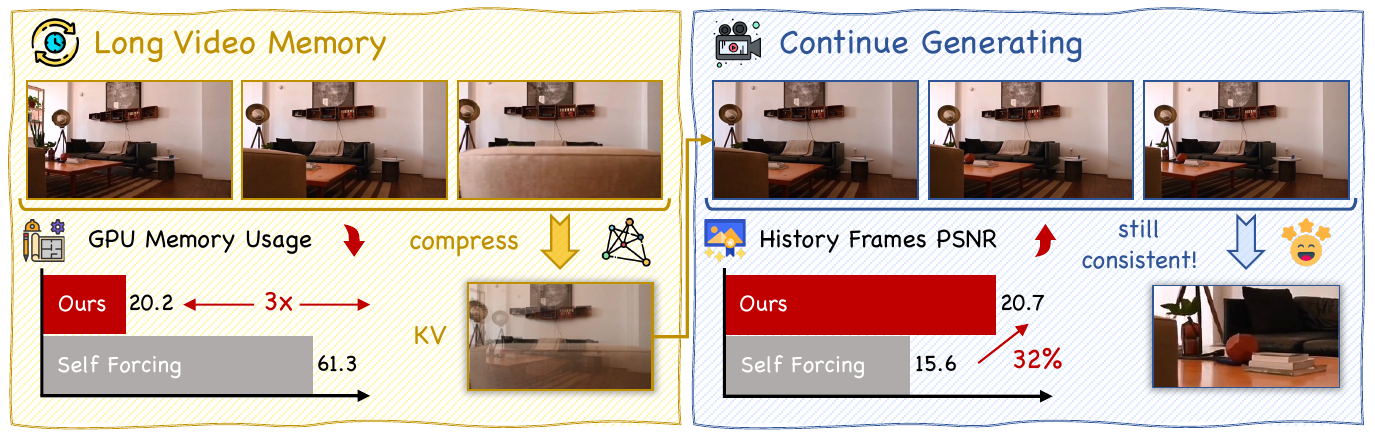}
  \caption{\textbf{The inference pipeline}. MAG performs real-time streaming video generation at 16 FPS on a single GPU. Compared to baselines, MAG achieves $3\times$ memory compression. Simultaneously, MAG is capable of generating scenes beyond the current field of view based on memory, ensuring global historical consistency.}
  \label{fig:teaser}
\end{figure}

\section{Introduction}
\label{sec:intro}

Research in the field of video generation is currently undergoing a paradigm shift from diffusion models based on bidirectional attention to autoregressive diffusion generation models operating in the temporal dimension. The former, exemplified by Wan2.1~\cite{wan2025}, is capable of generating high-fidelity short video clips from text prompts within a limited time interval. Furthermore, subsequent works have introduced additional control conditions~\cite{zhang2025flexiact,vace,bai2025recammaster}, allowing these models to generate diverse videos that better align with user requirements. However, the computational demands of bidirectional attention mean that users often face minutes of generation time to create video clips lasting only a few seconds. Moreover, constrained by fixed temporal windows, these diffusion models remain unable to generate minute-level videos. Consequently, frame-autoregressive models~\cite{zhang2025framepack,gu2025long, ai2025magi1autoregressivevideogeneration, jin2024pyramidal} have emerged as the most promising approach for efficient long video generation.

Frame-level autoregressive (frame-AR) video generation models are trained on spatial diffusion and can leverage temporal KV cache for acceleration. Compared to bidirectional attention, this causal attention mechanism effectively halves the computational load. Nevertheless, achieving high-fidelity frames typically requires setting the denoising steps between 20 and 50, which results in persistent high latency and low throughput during long video generation. For instance, MAGI-1~\cite{ai2025magi1autoregressivevideogeneration} requires several minutes to generate a single frame. To address this, CausVid~\cite{yin2025causvid} and Self Forcing~\cite{huang2025selfforcing} explored the use of DMD~\cite{yin2024improved, yin2024onestep} methods to convert lightweight bidirectional attention models into causal frame-AR models, successfully reducing the denoising steps to 4. Self-forcing~\cite{huang2025selfforcing} is capable of generating high-quality short video content in real-time on consumer GPUs, also achieving performance comparable to the original model. Recent work~\cite{yang2025longlive,liu2025rolling,cui2025self} has successfully transferred the Self Forcing~\cite{huang2025selfforcing} training methodology directly to long video tasks.

Despite these impressive advancements, existing long video generation methods~\cite{yang2025longlive,liu2025rolling,cui2025self,zhang2025framepack,ren2025videorag} continue to struggle with the trade-off between memory retention and computational resource consumption. Specifically, ensuring historical scene consistency—such as maintaining scene integrity when a camera returns to an original position after panning away—remains a critical challenge. Most recent works~\cite{yang2025longlive,liu2025rolling,cui2025self} have adopted window attention or rolling window operations, compromising the spatial complexity pressure inherent in long videos. For example, a one-second video clip can occupy nearly 1.5GB, meaning one minute of content is sufficient to fill the current state-of-the-art GPUs. However, reliance on short window attention leads to catastrophic forgetting; previous world models utilizing short windows~\cite{he2025matrix,yu2025gamefactory,zhang2025matrixgame,guo2025mineworld} often generate completely different scenes when looking left and then returning to the right. Later in this paper, we identify similar issues in recent video generation works, which we attribute to overly coarse memory strategy designs. Context as Memory~\cite{yu2025context} attempts to combine 3D perspectives with 2D representations to precisely select necessary frames, reducing redundant computation, but still necessitates retaining every historical frame since any frame could potentially be re-selected, leading to rapidly inflating memory consumption. Alternatively, TTT-video~\cite{dalal2025one} attempts to internalize memory into model parameters to avoid KV cache growth. However, optimizing these parameters during the inference sacrifices real-time performance, and the algorithm's complexity makes it difficult to scale up and deploy.

In this work, we analyze the training pipeline of recent approaches and identify a degenerate solution issue within the optimization objective. Consequently, we reformulate the training pipeline of Self Forcing~\cite{huang2025selfforcing} with historical context for long video generation, incorporating a loss function with text-free condition to reinforce the preservation of historical consistency. Furthermore, we introduce a learnable historical memory compression strategy into the frame-AR framework to simultaneously address the challenges of historical consistency and GPU memory consumption. We advocate decoupling memory compression and history-conditioned generation into two distinct tasks, necessitating the training of separate models. The memory model learns to reconstruct original pixel frames from the compressed KV cache, whereas the generator model learns to synthesize the next frame utilizing this compressed cache. Based on these insights, we propose \textbf{Memorize-and-Generate (MAG)}, a concise paradigm that not only precisely memorizes and reconstructs historical information but also achieves high compression ratios, significantly reducing memory usage without incurring additional computational overhead. Additionally, to facilitate fair comparison and evaluation of historical consistency across different methods, we curated \textbf{MAG-Bench}, a lightweight benchmark consisting entirely of videos with camera trajectories that leave and subsequently return to the scene, designed to accurately quantify a model's capacity for historical scene retention. Experimental results demonstrate that MAG exhibits superior consistency in video generation tasks while maintaining competitive performance in frame quality and text alignment metrics. On MAG-Bench, it demonstrates significantly better historical scene retention compared to existing methods.

\section{Related Work}
\label{sec:relate}
\noindent\textbf{Bidirectional Attention Video Generation.\quad} Both early research~\cite{ho2022video,he2022lvdm,chen2023videocrafter1,xing2023dynamicrafter} and recent high-quality short video generation~\cite{wan2025,kong2024hunyuanvideo,yang2024cogvideox} typically employ a bidirectional attention mechanism. This paradigm treats the video as a holistic entity, allowing the model to simultaneously perceive both past and future context during the generation process. Currently, mainstream architectures have evolved from early Spatio-temporal U-Nets~\cite{ronneberger2015u} to Diffusion Transformer (DiT)~\cite{peebles2023scalable} designs, enabling efficient scaling of model parameters and leading to the emergence of excellent open-source models~\cite{wan2025,kong2024hunyuanvideo,yang2024cogvideox,opensora,opensora2}. Notably, Wan2.1~\cite{wan2025} represents a state-of-the-art example; its lightweight 1.3B variant remains highly competitive even compared to significantly larger models. However, it is important to note that while bidirectional attention ensures high-quality context, it is inherently incompatible with the growing demand for real-time performance.\vspace{5pt}

\noindent\textbf{Autoregressive Video Generation.\quad} To transcend duration limits and enable streaming generation, research focus has gradually shifted toward temporal autoregressive paradigms. Initial autoregressive models~\cite{hong2022cogvideo,ren2025next,yan2021videogpt} directly discretized visual tokens via VQVAE~\cite{van2017neural} for full-sequence generation but yielded limited video quality. Subsequent works~\cite{kim2024fifo,jin2024pyramidal,li2024arlon} combined temporal autoregression with spatial diffusion, advancing frame-by-frame or chunk-by-chunk. However, these early explorations faced severe exposure bias, as errors accumulated along the temporal dimension, causing frame degradation. Diffusion Forcing~\cite{chen2024diffusion} proposed adding noise to each frame independently to enhance the model's error-correction capability. Similarly, Rolling Diffusion~\cite{kim24rolling} designed a window with progressively increasing noise for joint denoising and streaming output. Nevertheless, these methods failed to solve the exposure bias problem. Self Forcing~\cite{huang2025selfforcing} addressed this by enforcing strict training-inference consistency and utilizing the DMD distillation framework~\cite{yin2024onestep,yin2024improved} to produce high-quality content in real-time. Recent adaptations~\cite{yang2025longlive,cui2025self,liu2025rolling} have successfully transferred this method to long video tasks, generating minute-level videos without degradation. This demonstrates that by ensuring the training and inference processes are identical, the model can effectively manage temporal error accumulation autonomously.\\
\noindent\textbf{Memory Representations for Long-Term Consistency.\quad} Despite exciting progress in real-time performance and content quality, maintaining historical consistency remains a core challenge in long video generation. Furthermore, there is often an inherent trade-off between historical consistency and limited GPU memory. This issue is particularly critical for world models~\cite{he2025matrix,yu2025gamefactory,zhang2025matrixgame,guo2025mineworld}. Existing works~\cite{li2025vmem,huang2025memory,schneider2025worldexplorer,yu2025context} have primarily investigated three memory representation paradigms. 

\noindent\textbf{Explicit 3D Memory:} Representative methods like Memory Forcing~\cite{huang2025memory} and WorldExplorer~\cite{schneider2025worldexplorer} advocate converting video into 3D point cloud structures. These approaches progressively add new frames to a global 3D map and leverage reprojection as a geometric constraint to guide generation. Such explicit memory possesses a natural advantage in addressing spatial drift and historical consistency. However, their performance is heavily contingent on backend reconstruction algorithms~\cite{WangCKV0N25VGGT,video_depth_anything}. Failures in texture-poor or highly dynamic scenes can result in reconstruction errors that lead to persistent generation artifacts. 

\noindent\textbf{Implicit 2D Latent Memory:} Another line of work explores establishing memory within a 2D latent space. Genie 3~\cite{lin2024out,genie3} demonstrates that excellent historical consistency can be achieved using only 2D representations, though the technical details remain undisclosed. Context as Memory~\cite{yu2025context} designs a memory extraction strategy where the generation of the next frame relies solely on historical frames with significant viewpoint overlap. While this approach effectively combines 3D perspectives with 2D representations, the requirement to retain every historical frame continues to impose substantial hardware challenges. 

\noindent\textbf{Weight Memory:} TTT~\cite{sun2020test} proposes updating model weights during inference, effectively compressing information into the parameters. Theoretically, this method achieves a fixed $O(1)$ state size and infinite context memory, enabling precise reproduction of long-range details. TTT-video~\cite{dalal2025one} implements this approach, successfully generating animated shorts with preserved character consistency. However, performing optimization during the inference process introduces large computational burdens, thereby sacrificing real-time capabilities.

\section{Method}
\subsection{Preliminary}
\label{sec:preliminary}
Recent successful works~\cite{yang2025longlive,cui2025self,liu2025rolling} transfer the Self Forcing~\cite{huang2025selfforcing} to long video generation. The core method involves DMD distillation~\cite{yin2024onestep,yin2024improved}, which minimizes the KL divergence between the generator's output distribution $p_{\theta}^G(x)$ and the teacher model's output distribution $p^{\mathcal{T}}(x)$, utilizing short clips uniformly sampled from long videos for gradient backpropagation. For the sake of brevity and clarity, we omit the randomly sampled timestep $t$ in the following sections, removing a default expectation term to reduce formula length and complexity. The gradient of the optimization objective can be approximated as the difference between two score functions:

$$\begin{aligned}
\nabla_\theta\mathcal{L}_\mathrm{DMD} & =\mathbb{E}_x\left.\left[\nabla_\theta\operatorname{KL}\left(p_{\theta}^\mathcal{S}(x)\right.\Vert p^\mathcal{T}(x)\right)\right] \\
& \approx \left.\mathbb{E}_{\substack{i}\sim U\{1,k\}}\mathbb{E}_{\substack{z\sim \mathcal{N}(0,I)\\ x=G_\theta(z)} }\left[s^\mathcal{T}(x_i)-s_\theta^\mathcal{S}(x_i)\frac{dG_\theta(z_i)}{d\theta}\right.\right],
\end{aligned}$$

where $i \sim U\{1,k\}$, and $x_i$ and $z_i$ represent a clip uniformly sampled from $k$ segments used to calculate the DMD loss. Here, $z$ denotes Gaussian noise, $G_\theta(z)$ represents the output of the generator parameterized by $\theta$, $\mathcal{T}$ and $\mathcal{S}$ denote the teacher and student models respectively, and $s(x_i)$ is score function from teacher or student. During training, the student model utilizes Flow Matching~\cite{liu2022flow,lipman2022flow} to learn the generator's output; consequently, the student model distribution $p^\mathcal{S}$ can be viewed as a proxy distribution for the generator.
\subsection{Rethinking DMD Optimization in Long Video Generation}
\label{sec:rethink}
When $i > 1$ or the video extends to a longer duration, the generator's output distribution can be denoted as $p_{\theta}^G(x|h,T)$, where $h$ represents the history frames and $T$ represents the text condition. The model is required to generate new clips that are consistent with both the historical context and the text condition. However, the modeling approach described in~\cref{sec:preliminary}, which only relies on randomly sampled clips, neglects the critical role of historical information. In practice, recent works utilize the original T2V teacher model as a substitute for a more powerful teacher model capable of supporting history condition inputs. Consequently, the existing optimization objective is formulated as:
$$p_{\theta}^G(x|h,T)\rightarrow p^\mathcal{T}(x|T) \approx p^\mathcal{T}(x|h,T)$$
As the original teacher model lacks the ability to provide supervision signals for historical consistency, this implies the existence of a shortcut or \textbf{degenerate solution:} $p_{\theta}^G(x|h,T)\rightarrow p_{\theta}^G(x|T)$. First, text and historical context exhibit a high correlation, relying on text is sufficient to generate video of enough quality. Second, as the base model is a bidirectional T2V model, it inherently tends to converge towards relying on the text first when adapted to AR model, thereby neglecting the utilization of historical information. To address this degenerate solution, we introduce a simple modification: when $i > 1$, the generator predicts $x$ using an empty text condition. The form of the loss function remains unchanged, but the sampling source of $x$ is altered. The new loss function is defined as:
$$\begin{aligned}
\nabla_\theta\mathcal{L}_\mathrm{history} &=\mathbb{E}_{x\sim {p_{\theta}^G(x|h,\emptyset)}}\left.\left[\nabla_\theta D_{KL}\left(p_{\theta}^\mathcal{S}(x)\right.\Vert p^\mathcal{T}(x)\right)\right] \\
\nabla_\theta\mathcal{L} &= {(1-\lambda)} \nabla_\theta\mathcal{L}_\mathrm{DMD} + {\lambda} \nabla_\theta\mathcal{L}_\mathrm{history} 
\end{aligned}$$
Here, $\lambda$ is a hyperparameter balancing the two losses, specifically implemented via random sampling. During training, the model is forced to learn to generate next frame based solely on historical clips. This is a more challenging task, but it facilitates the learning and modeling of physical consistency and world knowledge. In contrast, the method in~\cref{sec:preliminary} tends to over-align with text information, ignoring the intrinsic contextual correlations within the autoregressive process.

\begin{figure}[h]
  \centering
  \includegraphics[width=122mm]{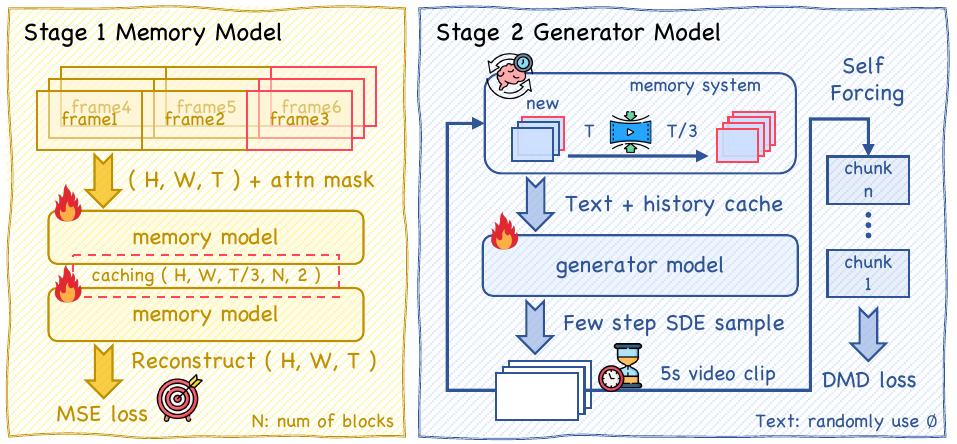}
  \caption{\textbf{The training pipeline}. The training process of MAG comprises two stages. In the first stage, we train the memory model for the triple compressed KV cache, retaining only one frame within a full attention block. The loss function requires the model to reconstruct the pixels of all frames in the block from the compressed cache. The process utilizes a customized attention mask to achieve efficient parallel training. In the second stage, we train the generator model within the long video DMD training framework to adapt to the compressed cache provided by the frozen memory model.
  }
  \label{fig:pipeline}
\end{figure}

\subsection{Memorize-and-Generate Framework}
Causal autoregressive models~\cite{yang2025longlive,cui2025self,huang2025selfforcing} first convert noise into video frames through a few denoising steps. Subsequently, these video frames are fed back into the model, where the KV cache of all tokens is retained block-by-block; this cache serves as the model's memory and historical condition. However, video sequences contain a massive number of tokens. Due to hardware constraints, it is impossible to retain all historical information. As mentioned in~\cref{sec:relate}, most works apply simple window attention\cite{yang2025longlive,cui2025self}, where the model only retains frames from the most recent 2-3 seconds, making it impossible to achieve historical consistency. To achieve this goal, we argue that all historical frame information should be preserved, as any fine-grained detail could potentially be reused. Therefore, to reduce the memory overhead associated with this goal, we propose decoupling the memory process of generating the final step's KV cache from the pipeline and implementing compression at the cache level. Next, we follow the methods of Self Forcing~\cite{huang2025selfforcing} and ~\cref{sec:rethink} to train the generator model. The frozen memory model executes the generation of the final step's KV cache, while the generator model maintains its original training objective. This scheme allows us to perform both near-lossless compression on long video content and real-time performance.

\subsection{Memory Model Design and Training}
We believe the design of the memory model should adhere to two principles: evaluable fidelity and no increase in inference latency. Ideally, it should exist directly in the form of a KV cache.

First, memory fidelity is the foundation of the final generation results; however, high-quality generation does not necessarily imply perfect memory preservation. For instance, methods based on sliding windows~\cite{yang2025longlive} or RNN hidden states~\cite{chen2025recurrent} are generally considered to suffer from information bottlenecks. However, these methods cannot quantify the extent of memory loss or identify the specific steps where the bottleneck occurs, thereby hindering our research and analysis of the relationship between memory and generation results.

Second, dynamic token compression during generation~\cite{zhang2025framepack} often requires regenerating the KV cache or extra computation, sacrificing real-time performance and design simplicity. We maintain that during streaming generation or input, once memory compression is completed, the historical cache or token sequence should not be frequently altered.

Drawing upon the aforementioned principles and inspired by Autoencoders (AE)~\cite{rumelhart1985learning}, we conceptualize the KV cache as compressed latent features and construct an encode-decode framework upon this foundation. In the streaming video generation process, the minimal output unit typically consists of several frames forming a block, within which full attention is applied. We treat the intra-block full attention computation as the \textbf{encoder}, retaining only a subset of the KV cache, such as the final frame. Subsequently, we task the model with denoising random noise to reconstruct all frames within the block based on this \begin{wrapfigure}{r}{0.5\textwidth} 
    \centering
    \includegraphics[width=0.9\linewidth]{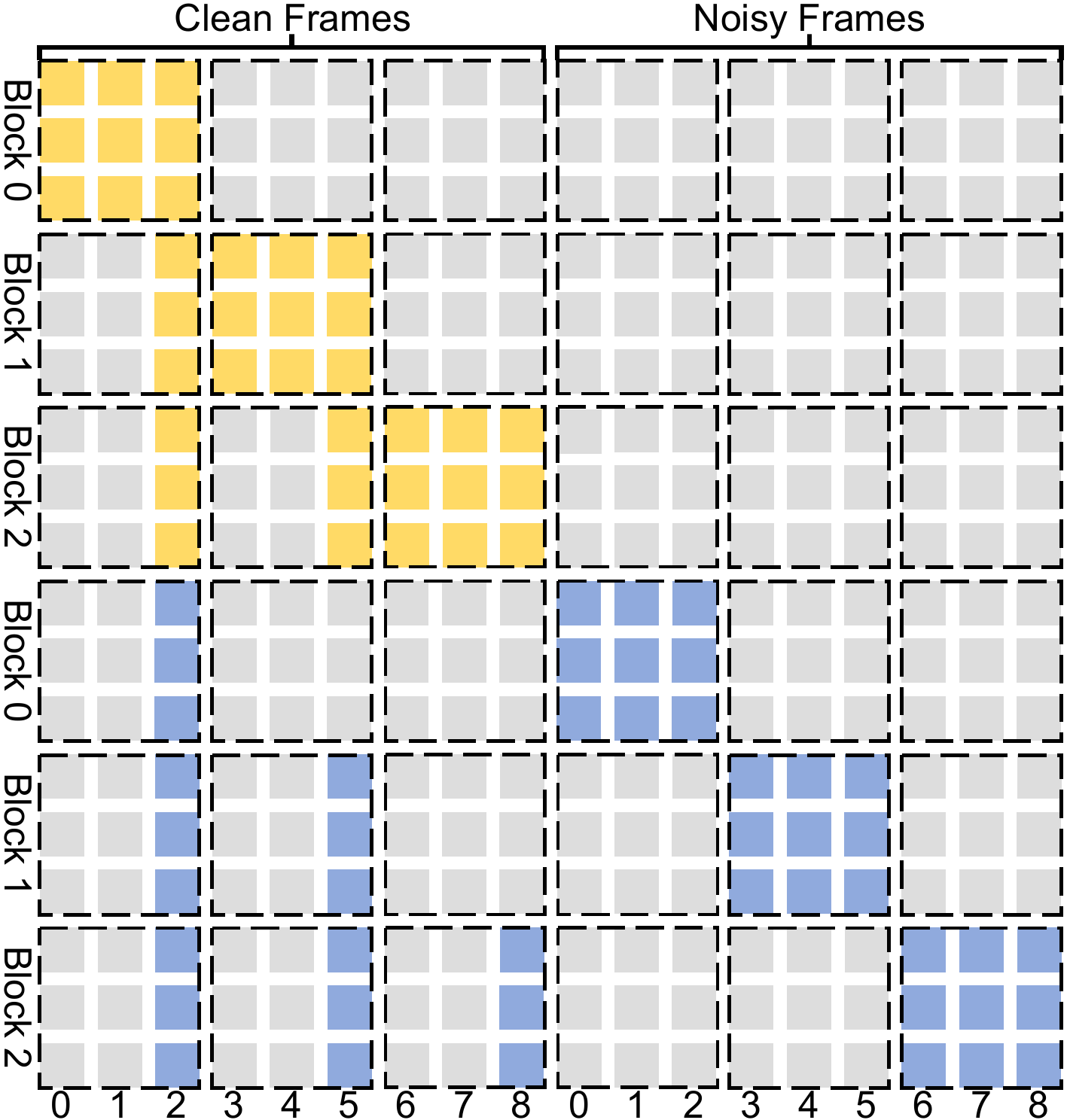}
    \caption{\textbf{The attention mask of memory model training}. We achieve efficient parallel training of the encode-decode process by concatenating noise and clean frame sequences. By masking out the KV cache of other frames within the block, the model is forced to compress information into the target cache.}
    \label{fig:attn_mask}
    \vspace{-20pt} %
\end{wrapfigure}
retained KV cache, functioning as the \textbf{decoder}. Since the dimensions of the KV cache represent a significant expansion relative to the input tokens, there is substantial compression at the cache level. This feasibility is illustrated on the left side of ~\cref{fig:pipeline}.

It is important to note that the encoder and decoder are implemented as a single model, the resulting memory model. Parameter sharing is enabled because the model can simultaneously produce the KV cache and the reconstruction results; these dual output branches naturally constitute an encode-decode workflow. We train with from few-step Flow Matching~\cite{liu2022flow,lipman2022flow}, which significantly enhances training efficiency. Driven by the gradients from the loss, the model learns during the encoding phase to compress the block's information into a single frame via full attention, while the decoding phase reconstructs the original pixels using this compressed information. Consequently, we can assess memory fidelity by simply observing the reconstructed video. Furthermore, to strictly ensure training-inference consistency, we randomize the start index of the Rotary Positional Embeddings~\cite{su2021roformer}. This encourages the model to learn that the compression task is independent of the video's temporal duration, allowing a memory model trained on short clips to be directly applied to long videos. Subsequent experiments demonstrate that this method achieves near-lossless compression. Moreover, the trained memory model can seamlessly replace the KV cache generation step in the final pipeline without introducing any additional computational overhead.

\subsection{Streaming Long Video Generation Training}

The training of the generator model in MAG primarily follows the protocols established in LongLive~\cite{yang2025longlive} and Self-forcing++~\cite{cui2025self}. However, distinct from these works, we employ the modeling approach detailed in~\cref{sec:rethink}. In each training step, the generator produces a 5-second short video clip, which may be conditioned on an empty text prompt. Subsequently, we calculate the DMD loss based on this clip to obtain the supervision signal. When the generator operates with an empty text condition, it is compelled to generate correct content based solely on historical information to align with the signal produced by the teacher model, thereby reinforcing the constraint of the historical condition. We then proceed with several identical generation steps and the resulting short video clips are utilized to update the student model, ensuring it represents the current output distribution of the generator. In practice, by calibrating the ratio of generator training steps to student model training steps, as well as the number of short clips within the long video, we ensure that during the rolling process, the generator's training provides uniform supervision across every time step. This guarantees the model's robustness over long-duration generation.

\begin{figure}[h]
  \centering
  \includegraphics[width=122mm]{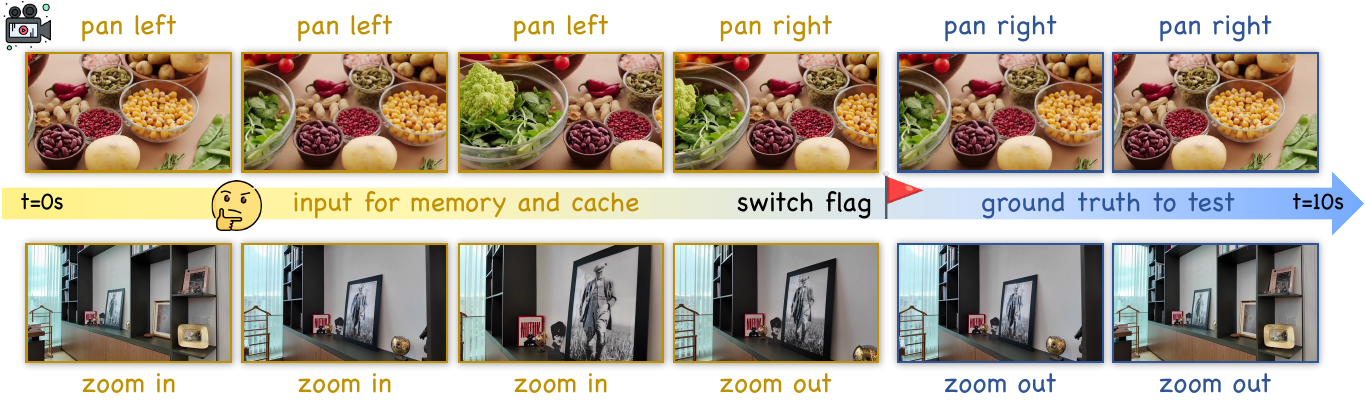}
  \caption{\textbf{Examples from MAG-Bench.} MAG-Bench is a lightweight benchmark comprising 176 videos featuring indoor, outdoor, object, and video game scenes. The benchmark also provides appropriate switch times to guide the model toward correct continuation using a few frames. 
  }
  \label{fig:mag_bench_example}
  \vspace{-30pt}
\end{figure}

\subsection{MAG-Bench: Historical Consistency Evaluation}

Historical consistency is particularly critical for future video generation tasks involving complex camera movements or even scene cuts. However, existing benchmarks primarily focus on image quality and text alignment, lacking dedicated data with back-and-forth camera movements to evaluate the model's ability to retain objects that have exited the frame. To bridge this gap, we collected a lightweight dataset. Furthermore, we discovered that simply inputting historical frames with active camera movement into the KV cache effectively guides the model to continue generating scenes that have moved out of frame, thereby facilitating evaluation. Since this process adheres to a "memorize-and then-generate" workflow, we term it \textbf{MAG-Bench}. To ensure that the camera movement videos in MAG-Bench are strictly symmetrical to calculate the reconstruction loss, we first collected high-quality videos with singular camera movements. We then synthesized high-quality "scene backtracking" videos through reverse playback. A schematic of the videos within the benchmark, along with the partition between memory and generation segments, is illustrated in~\cref{fig:mag_bench_example}.

\section{Experiments}
\subsection{Implementation Details}
\noindent\textbf{Training.} Following concurrent work, we utilize Wan2.1-T2V-1.3B~\cite{wan2025} as our base model to ensure fair comparison; this model generates 5-second clips at 16 FPS with a resolution of $832 \times 480$. First, adhering to the Self Forcing~\cite{huang2025selfforcing} training pipeline, we train the ODE-initialized model for 300 steps. This reproduces the capabilities of Self Forcing,~\cite{huang2025selfforcing} enabling the generation of 5-second short videos in few-step conditions. Subsequently, we proceed to train the memory model and generator model according to the scheme outlined in ~\cref{sec:rethink}. The memory model is initialized from the aforementioned 300-step Self Forcing model and trained for 2,000 steps on VPData~\cite{bian2025videopainter}, which contains 390K high-quality real-world videos, using empty text as the condition. The generator model is then initialized from the trained memory model. This strategy ensures that the feature space of the cache is shared during the early stages of training, thereby stabilizing the subsequent process. Following concurrent work, the generator model is trained for approximately 1,400 steps using text prompts sampled from VidProM~\cite{wang2024vidprom}, which have been extended by LLM. For all training phases, whether the supervised training of the memory model or the DMD training of the generator model, we set the batch size to 64, the generator learning rate to $2.0 \times 10^{-6}$, and the student model learning rate to $4.0 \times 10^{-7}$, consistent with or similar to concurrent studies~\cite{yang2025longlive,huang2025selfforcing}. Furthermore, during generator training, we set the probability of using empty text conditioning to 0.6. Each text prompt generates a long video consisting of 7 clips. To ensure the generator receives uniform supervision across the temporal dimension during the rolling process, we adopt a strategy where the student model is trained on 5 clips for every 1 clip used to train the generator model.

\begin{figure}[h]
  \centering
  \vspace{-15pt}
  \includegraphics[width=122mm]{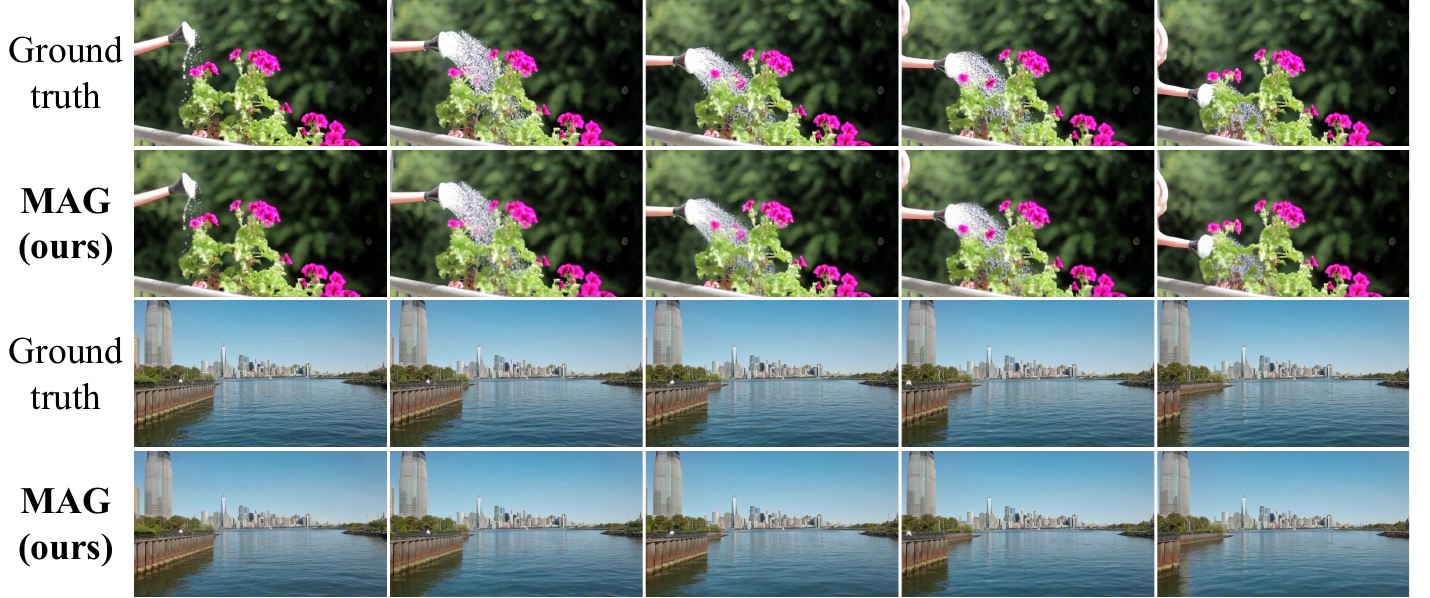}
  \caption{\textbf{Visualization of Memory Model reconstruction results.} We display two examples featuring texture detail variations and significant camera movement. Visually, the trained Memory Model achieves near-lossless reconstruction of the original pixels under a $3\times$ compression setting.
  }
  \label{fig:vpdata}
  \vspace{-15pt}
\end{figure}

\noindent\textbf{Evaluation.} For the memory model, we utilize the standard training and test splits of VPData~\cite{bian2025videopainter} and report PSNR, SSIM, LPIPS, and MSE on the test set to evaluate the effectiveness of compression and reconstruction. For the generator model, we employ VBench~\cite{huang2023vbench} and VBench-Long
~\cite{zheng2025vbench2}, to assess performance in text-to-video tasks, covering both quality and text alignment. It is important to note that, to ensure fair comparison with recent work, we follow Self Forcing~\cite{huang2025selfforcing} by using extended prompts for both the 5-second tests and the 30-second tests. For MAG-Bench, we refer~\cite{yu2025context} and report the PSNR, SSIM, and LPIPS of the predicted video compared to the ground truth to quantify memory capability on pixel level. During experiments, we observed that slight discrepancies in camera movement speed between the predicted video and ground truth could lead to significant pixel-level errors, even if the scene consistency was perceptually accurate. Therefore, we report metrics based on Best Match LPIPS: we first match each predicted frame to the most similar ground truth frame based on the perceptual loss of the pretrained model, and then calculate the remaining metrics. All generation results from distilled models are sampled using identical initial noise and random seeds, utilizing the parameter settings provided in the corresponding papers and codebases. FPS tests on a H100 GPU.
\begin{figure}[h]
\vspace{-10pt}
  \centering
  \includegraphics[width=122mm]{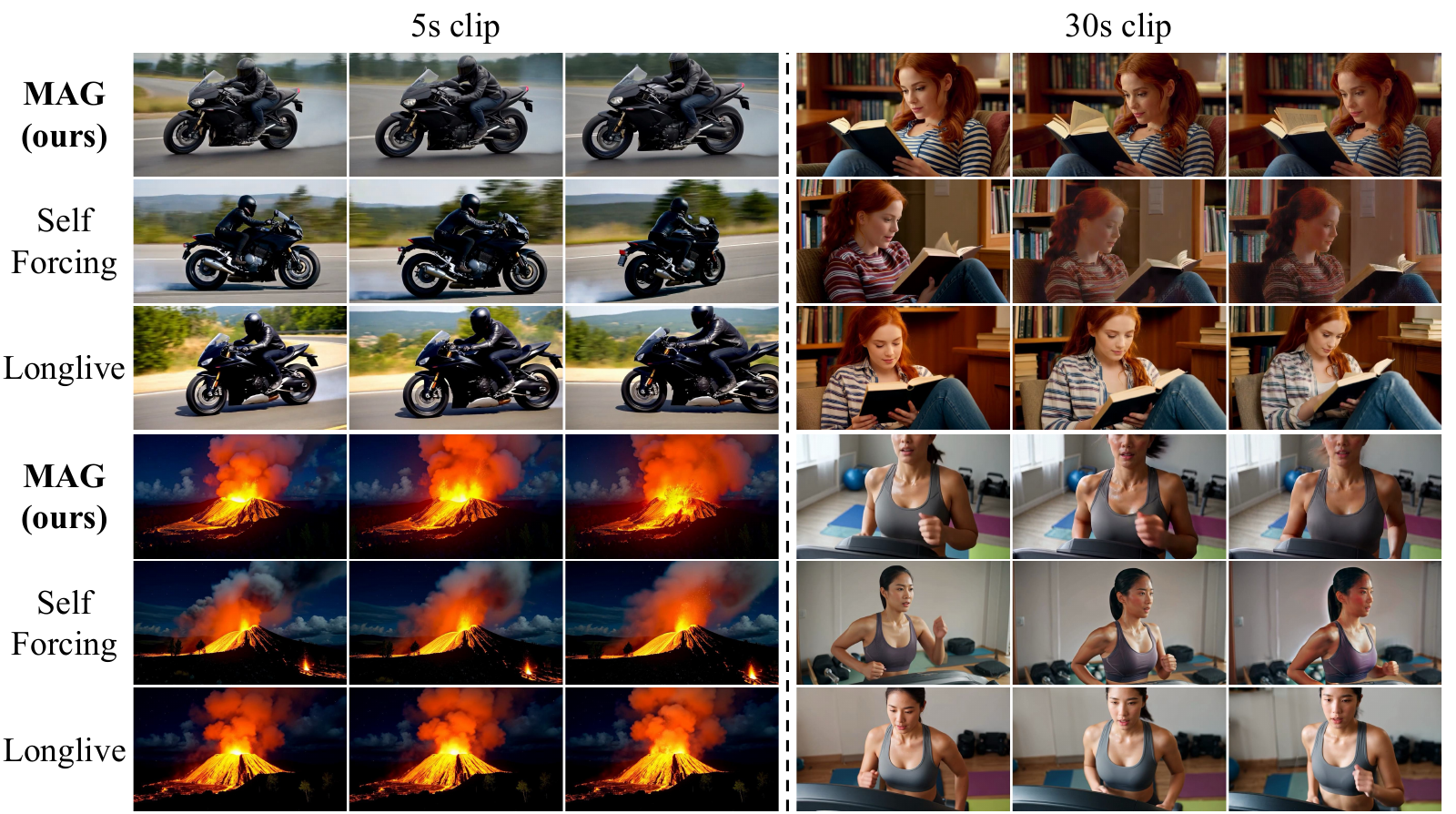}
  \caption{\textbf{Qualitative comparison on T2V tasks}. We present 5-second and 30-second video clips sampled from VBench~\cite{huang2023vbench} and VBench-Long~\cite{zheng2025vbench2}, respectively. All methods utilize identical prompts and random initialization noise.
  }
  \label{fig:vbench}
  \vspace{-30pt}
\end{figure}

\begin{table}[h]
\caption{
\textbf{Quantitative comparison with relevant baselines on the 5-second VBench~\cite{huang2023vbench}}. We compare against recent and representative open-source methods with similar parameter sizes and distillation processes. Evaluations are conducted using extended VBench prompts. "–" denotes that the data is cited from the reference but this metric was not disclosed. FPS is measured on a single H100 GPU.}
  \label{tab:short}
  \centering
  \setlength\tabcolsep{4pt}
\resizebox{\linewidth}{!}{
\begin{tabular}{lcccccc}
\toprule
  \multirow{2}{*}{Model}& \multirow{2}{*}{\makecell{Throughput\\FPS$\uparrow$}}& \multicolumn{5}{c}{Vbench scores on 5s $\uparrow$}\\
  & & \multirow{1}{*}{Total}& \multirow{1}{*}{Quality}& \multirow{1}{*}{Semantic}& \multirow{1}{*}{Background}& \multirow{1}{*}{Subject}\\
  \midrule
  \rowcolor{AliceBlue}
  \multicolumn{7}{l}{\textit{Multi-step model}}\\
  SkyReels-V2~\cite{chen2025skyreelsv2infinitelengthfilmgenerative} & 0.49 & 82.67 & 84.70 & 74.53 & - & -\\
  Wan2.1~\cite{wan2025} & 0.78 & 84.26 & 85.30 & 80.09 & 97.29 & 96.34\\
  \midrule
  \rowcolor{AliceBlue}
  \multicolumn{7}{l}{\textit{Few-step distillation model}}\\
  CausVid~\cite{yin2025causvid} & 17.0 & 82.46 & 83.61 & 77.84 & - & -\\
  Self Forcing~\cite{huang2025selfforcing} & 17.0 & \textbf{83.98} & \textbf{84.75} & 80.86 & 96.21 & 96.80\\
  Self Forcing++~\cite{cui2025self} & 17.0 & 83.11 & 83.79 & 80.37 & - & -\\
  Longlive~\cite{yang2025longlive} & 20.7 & 83.32 & 83.99 & 80.68 & 96.41 & 96.54\\
  \midrule
  MAG & \textbf{21.7} & 83.52 & 84.11 & \textbf{81.14} & \textbf{97.44} & \textbf{97.02}\\
  \bottomrule
\end{tabular}
}
\end{table}

\begin{table}[h]
\caption{
\textbf{Quantitative comparison with relevant baselines on the 30-second VBench-long~\cite{zheng2025vbench2}}. We compare against recent long video generation methods based on Self Forcing. Evaluations are conducted using extended VBench prompts.}
  \label{tab:long}
  \centering
  \setlength\tabcolsep{7pt}
\begin{tabular}{l|ccccc}
\toprule
  \multirow{2}{*}{Model}& \multicolumn{5}{c}{Vbench scores on 30s $\uparrow$}\\
  &\multirow{1}{*}{Total} & \multirow{1}{*}{Quality}& \multirow{1}{*}{Semantic}& \multirow{1}{*}{Background}& \multirow{1}{*}{Subject}\\
  \midrule
  Self Forcing~\cite{huang2025selfforcing} & 82.57 & 83.30 & 79.68 & 97.03 & 97.80\\
  Longlive~\cite{yang2025longlive} & 82.69 & 83.28 & 80.32 & 97.21 & 98.36\\
  \midrule
  MAG & \textbf{82.85} & \textbf{83.30} & \textbf{81.04} & \textbf{97.99} & \textbf{99.18}\\
  \bottomrule
\end{tabular}
\vspace{-10pt}
\end{table}

\subsection{Text-to-Video Generation Comparison}
We selected representative recent models as baselines. Wan2.1~\cite{wan2025} serves as our base model and stands as an excellent example of open-source bidirectional attention models for short video generation. SkyReels-V2~\cite{chen2025skyreelsv2infinitelengthfilmgenerative}, a diffusion forcing model of the same size, represents non-distilled autoregressive generation. Self Forcing~\cite{huang2025selfforcing} and LongLive~\cite{yang2025longlive} represent significant breakthroughs in recent distillation work; Self-forcing uses full history as a condition to generate 5-second clips, while LongLive~\cite{yang2025longlive} employs a 6-frame sliding window attention to generate minute-level videos. Note that Wan2.1~\cite{wan2025} and SkyReels-V2~\cite{chen2025skyreelsv2infinitelengthfilmgenerative} require dozens of denoising steps to achieve high-quality video, which naturally results in higher image quality scores. We primarily compare our approach against distilled works with consistent experimental settings to ensure fair evaluation. As shown in ~\cref{tab:short} and ~\cref{fig:vbench}, our model achieves a highly competitive score of 83.52 on the short video generation task. Furthermore, our method outperforms existing approaches in both background and object consistency, attributed to our maintenance of cache fidelity and a modeling approach that explicitly prioritizes historical consistency. Similar results are observed in long video generation tasks. Moreover, our model achieves a real-time inference speed of 21.7 FPS, which is the fastest among the compared methods. This speed advantage stems from our denser historical information compression, which reduces the sequence length required for attention. Although LongLive~\cite{yang2025longlive} utilizes a smaller window, its use of LoRA~\cite{hu2022lora} adapters and the sliding window shift operations introduces computational overhead; consequently, our method maintains a slightly faster average speed than other methods.

\subsection{Historical Consistency Comparison}
\begin{figure}[h]
\vspace{-10pt}
  \centering
  \includegraphics[width=122mm]{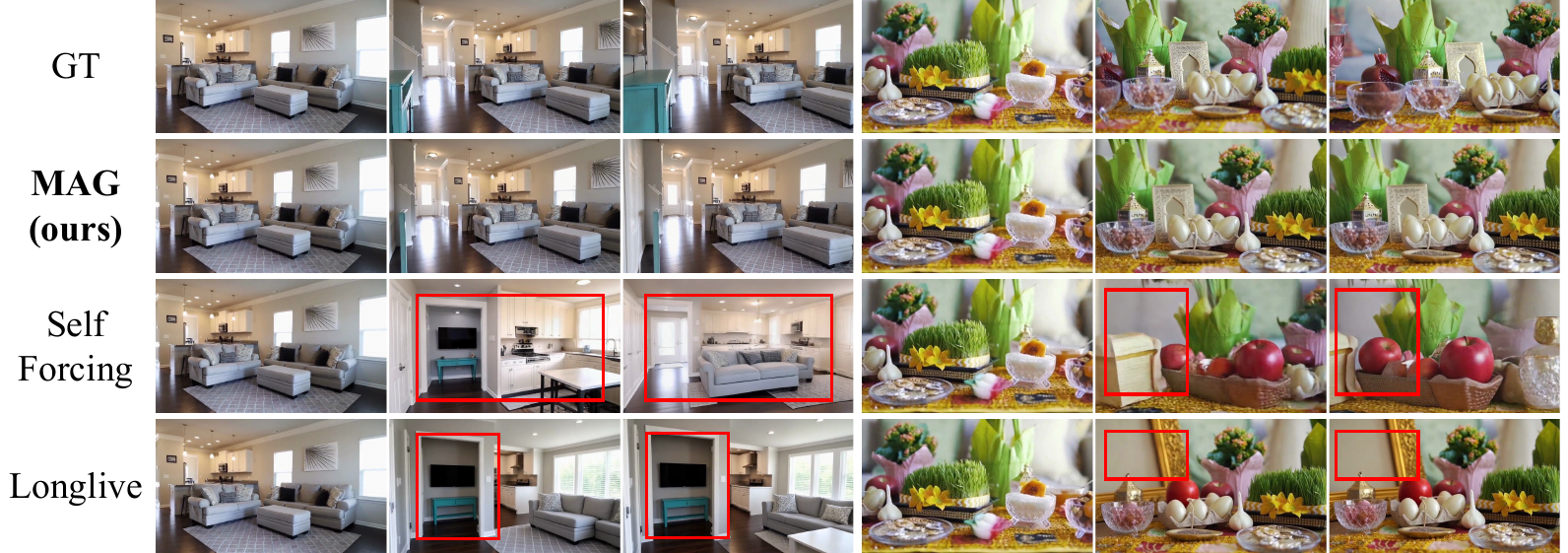}
  \caption{\textbf{Qualitative comparison on MAG-Bench.} We primarily display the visual results of comparable distilled models. Prior to these frames, the models receive and memorize historical frames. Red boxes highlight instances of scene forgetting and hallucinations exhibited by other methods.
  }
  \label{fig:mag_bench_compare}
\vspace{-30pt}
\end{figure}

We selected recent distillation methods for a comparative analysis of historical consistency. As illustrated in ~\cref{tab:mag}, our method significantly outperforms existing approaches across quantitative metrics. This is primarily because our method retains all historical information, whereas sliding window operations typically preserve only the most recent 2-3 seconds of history. Furthermore, while Self Forcing~\cite{huang2025selfforcing} and CausaVid~\cite{yin2025causvid} also retain the full historical frame cache, they do not enforce the model to learn the utilization of historical information during training. ~\cref{fig:mag_bench_compare} demonstrates a scenario where the camera leaves a scene and subsequently returns via camera movement guidance. Only our method maintains the best scene consistency, while other methods exhibit forgetting and hallucinations in various regions.

\begin{table}[h]
\caption{\textbf{Quantitative experiments on our MAG-Bench}. "Ground Truth" and "History Context" denote the model continuing generation based on ground truth frames and its own predicted frames, respectively, with the latter representing a significantly more challenging task.}
  \label{tab:mag}
  \centering
  \setlength\tabcolsep{7pt}
\begin{tabular}{l|ccc|ccc}
\toprule
  \multirow{2}{*}{Method}& \multicolumn{3}{c|}{History Context Comparison} & \multicolumn{3}{c}{Ground Truth Comparison}\\
& \multirow{1}{*}{PSNR$\uparrow$}& \multirow{1}{*}{SSIM$\uparrow$}& \multirow{1}{*}{LPIPS$\downarrow$}& \multirow{1}{*}{PSNR$\uparrow$}& \multirow{1}{*}{SSIM$\uparrow$}& \multirow{1}{*}{LPIPS$\downarrow$}\\
  \midrule
  Self Forcing~\cite{huang2025selfforcing} & 14.46 & 0.48 & 0.49 & 15.65 & 0.51 & 0.42\\
  CausVid~\cite{yin2025causvid} & 15.13 & 0.50 & 0.41 & 17.21 & 0.56 & 0.31\\
  Longlive~\cite{yang2025longlive} & 16.42 & 0.53 & 0.32 & 18.92 & 0.62 & 0.22\\
  \midrule
  w/o stage 1 & 17.19 & 0.54 & 0.31 & 19.04 & 0.60 & 0.22\\
  MAG & \textbf{18.99} & \textbf{0.60} & \textbf{0.23} & \textbf{20.77} & \textbf{0.66} & \textbf{0.17}\\
  \bottomrule
\end{tabular}
\vspace{-30pt}
\end{table}

\subsection{Ablation Studies}
\noindent\textbf{Memory Model Compression Rate.\quad} We adopt a strategy of retaining only the cache of the final frame for subsequent use after performing full attention computation on a block of frames. In our setting, the compression rate is equivalent to the number of frames within a block, which acts as a hyperparameter affecting throughput in other works.
~\cref{tab:compress} indicates that reconstruction quality decreases slightly as the compression rate increases. ~\cref{fig:vpdata} shows that when the compression rate is set to 3, the model reconstructs original pixels with visually negligible impact. Since a block size of three frames is a widely accepted parameter balancing throughput and latency~\cite{cui2025self,huang2025selfforcing,yang2025longlive}, we selected a compression rate of 3. However, the results in ~\cref{tab:compress} suggest that higher compression rates remain a viable avenue for future exploration.
\begin{table}[!h]
\vspace{-10pt}
\caption{
\textbf{Ablation study on compression rates}. We vary the compression rates by adjusting the number of frames contained within a block. "Block=1" indicates no compression.}
  \label{tab:compress}
  \centering
  \setlength\tabcolsep{7pt}
\begin{tabular}{l|cccc}
\toprule
  \multirow{1}{*}{Rates}& \multirow{1}{*}{PSNR$\uparrow$}& \multirow{1}{*}{SSIM$\uparrow$}& \multirow{1}{*}{LPIPS$\downarrow$}& \multirow{1}{*}{MSE$_{\times10^2}\downarrow$}\\
  \midrule
  block=1 & 34.81 & 0.93 & 0.025 & 0.08 \\
  block=3 & 31.73 & 0.90 & 0.045 & 0.56 \\
  block=4 & 29.89 & 0.88 & 0.059 & 1.28 \\
  block=5 & 28.64 & 0.86 & 0.071 & 1.96 \\
  \bottomrule
\end{tabular}
\vspace{-10pt}
\end{table}

\noindent\textbf{Memory Model Ablation Study.\quad} To demonstrate the necessity of memory compression, we trained a baseline method that employs direct $3\times$ downsampling. As shown in ~\cref{tab:mag}, omitting the first stage of memory compression training results in poorer consistency. Simple downsampling discards a substantial amount of detail and information, leading to potential forgetting during scene reconstruction. Therefore, ensuring cache fidelity through the first stage of training proves to be an effective strategy.

\section{Discussion}
\noindent\textbf{Conclusion.} In this work, we propose MAG, a framework for long video generation comprising two models dedicated to memory compression and next-frame generation, respectively. To evaluate the historical scene consistency of existing methods, we construct MAG-Bench, a lightweight benchmark. Experimental results demonstrate that the memory model can reconstruct original pixels under $3\times$ compression. Furthermore, the generator model synthesizes high-quality content in real-time with superior background and subject consistency. Besides, it significantly outperforms existing methods in maintaining historical consistency.\vspace{5pt}

\noindent\textbf{Limitations and Future Work.} During our experiments, we identified two primary limitations. (1) While this work ensures the fidelity of the compressed KV cache, the lack of data explicitly targeting context consistency makes it difficult for the generator model to learn how to optimally select and utilize extensive historical frames. (2) Although the DMD distillation framework is data-free, this characteristic hinders its direct extension to action-based world models. Substantial resources are still required to train a capable teacher model. In future work, we aim to address these challenges to enhance the feasibility of realizing world models.

\bibliographystyle{splncs04}
\bibliography{main}
\end{document}